\documentclass[english,5p]{elsarticle}
\usepackage[T1]{fontenc}
\usepackage[latin9]{inputenc}
\usepackage{color}
\usepackage{babel}
\usepackage{array}
\usepackage{graphicx}
\usepackage[unicode=true,
 bookmarks=false,
 breaklinks=false,pdfborder={0 0 0},backref=false,colorlinks=true]
 {hyperref}
\hypersetup{pdftitle={Few-shot learning using pre-training and shots, enriched by pre-trained samples},
 pdfauthor={Detlef Schmicker}}

\makeatletter

\AtBeginDocument{\providecommand\tabref[1]{\ref{tab:#1}}}
\providecommand{\tabularnewline}{\\}

\makeatother

\begin{document}

\begin{frontmatter}{}

\title{Few-shot learning using pre-training and shots, enriched by pre-trained
samples}

\author{Detlef Schmicker}

\address{Ludgeriplatz 31, 47057 Duisburg, Germany}

\ead{dschmicker@physik.de}
\begin{abstract}
We use the EMNIST dataset of handwritten digits to test a simple approach
for few-shot learning. A fully connected neural network is pre-trained
with a subset of the 10 digits and used for few-shot learning with
untrained digits. Two basic ideas are introduced: during few-shot
learning the learning of the first layer is disabled, and for every
shot a previously unknown digit is used together with four previously
trained digits for the gradient descend, until a predefined threshold
condition is fulfilled. This way we reach about 90\% accuracy after
10 shots.\end{abstract}

\end{frontmatter}{}

\section{Introduction}

Neural networks have shown stunning success \cite{alphago}, but training
requires a big database or the ability to generate a high amount of
training samples from rules \cite{alphago_zero}. Obviously it is
not always possible to fulfill this, for example if you want to teach
a computer manually. Therefore we are interested to learn from few
examples. A number of quite different approaches were proposed and
it is even difficult to classify our approach within the discussed
ones \cite{few_shot_review}. Our approach could be classified as
a refinement of existing parameters, as well as embedding learning,
no meta learning and one could argue that we use external memory,
as we use samples from pre-training during few-shot learning. 

We combine a number of well known concepts. To avoid optimizing the
neural network for a specific task we use only fully connected layers
with weights and without bias. The activation function has an output
between 0 and 1 to keep it similar to human neurons (figure \ref{fig:The-activation-function}).
The function value is close to 0 for input 0, similar to human neurons,
where a 0 (not firing) has no effect on the connected neuron, as a
zero output is only multiplied by a weight. Some tests indicate, that
this restriction is not very important, but as it similar to human
neurons, we keep it that way. This choice might also be useful, if
one wants to add local plasticity \cite{BALDI201651}. In this work
we use ``plasticity'' only by reducing the learning rate of the
first layer.

The samples are taken from the EMNIST dataset of handwritten digits
\cite{cohen2017emnist}. The EMNIST dataset contains 280000 digits
from more than 500 different writers, classified using 10 labels,
representing the digits. The subset belonging to 8 of the 10 digits
are used for pre-training. The motivation is, that humans have seen
a lot of lines and shapes during there live before they try to read
digits. Therefore they have a pre-trained brain. After pre-training
the neural network learns the two remaining digits, which it has never
seen before, from few examples with a few shot procedure, we will
describe in detail. 

The pre-training is done with a standard gradient descend. The few-shot
learning is also done this way, but only uses one new sample at a
time, and stops learning depending on a stop-criteria. Without any
additional measures, this approach fails. Two measures were necessary
to succeed: 

the learning for the first layer has to be disabled, or at least slowed
down, and 

with every shot some previously known samples have to be added. 

This helps the network remembering the old labels, similar to human
learning: if you do not use old knowledge, you forget it. This way
we reach about 90\% accuracy with 10 shot learning. The jupyter notebook
with the calculations is available \cite{Kluyver:2016aa,SMIC2020}.

\section{The neural network}

\begin{figure}
\begin{centering}
\includegraphics[width=6cm]{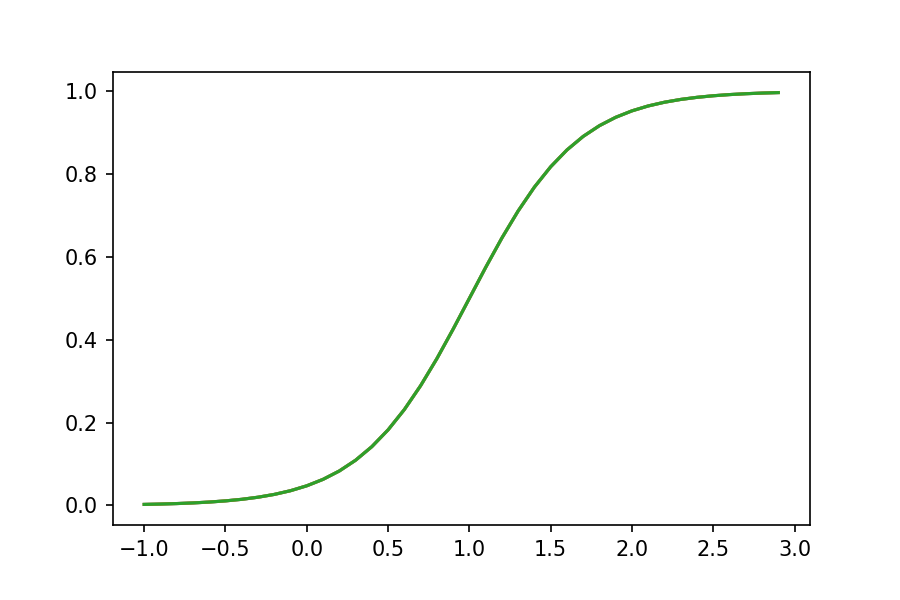}
\par\end{centering}

\caption{\label{fig:The-activation-function}The activation function used.
Similar to human neurons the output is close to zero, if there is
no input from the previous layer.}
\end{figure}

\begin{figure}
\begin{centering}
\includegraphics{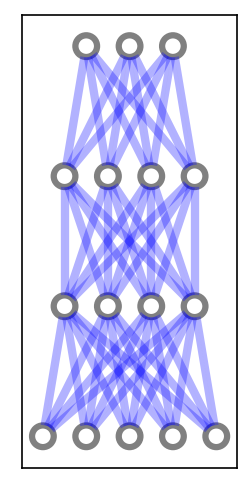}
\par\end{centering}

\caption{\label{fig:The-net-used}This graph shows a fully connected neural
network, with 5 inputs, two hidden layers of size 4 and 3 outputs.
Throughout this work we use this neural network with larger layers.}
\end{figure}

Each fully connected layer has $i$ inputs $x_{i}$ and $j$ outputs
$y_{j}$, which are related by $y_{j}=\sum_{i}w_{ij}x_{i}$. Each
activation layer takes $k$ inputs $x_{k}$ and has $k$ outputs $y_{k}$
with the relation $y_{k}=\frac{1}{1-e^{-(3(x_{k}-1))^{2}}}$, which
is a sigmoid function, scaled and shifted in x-direction (figure \ref{fig:The-activation-function}).
The neural network has one input layer, two hidden layers and an output
layer as in figure \ref{fig:The-net-used}, 
but the size of the layers are larger in the following. The loss function
$l(y_{i})=\frac{1}{2}\sum_{i}(y_{i}-\hat{y}_{i})^{2}$ is used, with
$\hat{y}$ being the target value from the training data. Every output
of the network corresponds to one of the labels (digits), with this
target being 1 and all others 0. The gray scale inputs are scaled
to the range between 0 and 1. Usually the weights are random initialized
with a mean value of 0 and a standard derivative of 0.1 and clipped
during training to a value between -1 and 1, similar to human neurons
not firing with infinite intensity. The minimum of the loss function
with respect to the weights is calculated using a gradient descend
method implemented in python \cite{SMIC2020}.

\section{The pre-training procedure}

From the EMNIST dataset of handwritten digits we chose 8 digits for
the pre-training. With them a neural network with 28x28 input pixels,
two hidden layers of size 64 and 10 outputs is trained. From the 10
outputs only 8 are used for pre-training, the other two are reserved
for few-shot learning of the remaining two digits (the digits 2 and
8 in this work). We train for 100000 batches with a batch size of
1000. The EMNIST dataset of handwritten digits contains 280000 digits,
which are split into 240000 digits for training and 40000 digits for
testing. From the 240000 digits only 192000 digits belong to the 8
digits we use for pre-training. The pre-training procedure results
in an accuracy of about 98\%, measured with 1000 images from the testing
dataset. A prediction is considered correct, if the output belonging
to the correct digit has the highest value of all outputs.

\section{\label{sec:The-few-shot}The few-shot procedure}

Two modifications as compared to the pre-training are used during
few-shot learning:
\begin{enumerate}
\item learning for the first layer is disabled \emph{and} 
\item for every shot one new sample of the previously unknown digits is
combined with 4 samples, which where part of pre-training. 
\end{enumerate}
With this mini-batch a gradient descend is performed until the two
following stop criteria for the new sample are fulfilled:
\begin{enumerate}
\item the value of the output, corresponding to the correct label, is bigger
than 0.3, \emph{and}
\item the value of the output, corresponding to the correct label, is more
than 1.5 times the value of the second highest output.
\end{enumerate}
The samples used during pre-training are not taken into account for
the stop criteria. For every shot we perform a gradient descend with
two new samples, belonging to both labels not used during pre-training.

\section{Experimental results}

\begin{table}
\begin{centering}
\begin{tabular}{|c|>{\centering}p{1.5cm}|>{\centering}p{1.5cm}|>{\centering}p{1.5cm}|>{\centering}p{1.5cm}|}
\hline 
shot & accuracy old digits & accuracy new digits & accuracy only new digits & overall accuracy\tabularnewline
\hline 
\hline 
1 & 0.895 & 0.492 & 0.776 & 0.815\tabularnewline
\hline 
2 & 0.952 & 0.574 & 0.754 & 0.879\tabularnewline
\hline 
3 & 0.950 & 0.521 & 0.667 & 0.870\tabularnewline
\hline 
4 & 0.960 & 0.535 & 0.730 & 0.878\tabularnewline
\hline 
5 & 0.856 & 0.621 & 0.677 & 0.818\tabularnewline
\hline 
6 & 0.961 & 0.708 & 0.839 & 0.905\tabularnewline
\hline 
7 & 0.958 & 0.666 & 0.806 & 0.900\tabularnewline
\hline 
8 & 0.949 & 0.747 & 0.844 & 0.911\tabularnewline
\hline 
9 & 0.954 & 0.811 & 0.918 & 0.923\tabularnewline
\hline 
10 & 0.926 & 0.836 & 0.919 & 0.908\tabularnewline
\hline 
\end{tabular}
\par\end{centering}

\caption{\label{tab:Experimental-results-few}Experimental results for few-shot
learning. The pre-training is done with 100000 batches of size 1000.
Learning of the first layer is disabled during few-shot learning,
and every shot is enriched by four samples, known from pre-training.}
\end{table}

The result of the few-shot learning, measured using the testing data
set, is shown in table \ref{tab:Experimental-results-few}. The first
accuracy column reports the accuracy, taking only pre-trained digits
into account, the second reports how good the new digits are recognized.
The ``accuracy only new digits'' column reports the accuracy, if
one tries to distinguish the new digits. Therefore it is just verified
that the neurons corresponding to the additional digits show the higher
value. The last column measures how good the neural network recognizes
all 10 digits. A accuracy of around 90\% seems quite convincing for
few-shot learning of handwritten digits from many different writers.

We check both modifications from section \ref{sec:The-few-shot} experimentally:
first we do not disable learning for the first layer (table \ref{tab:Experimental-results-without-fixed})
and second we do not enrich the few-shot sample with already learned
samples (table \ref{tab:Experimental-results-without-enriched}).
In both cases few-shot learning fails, as the neural network forgets
the pre-trained samples, indicating that both ideas are important
for the success. 
\begin{table}
\begin{centering}
\begin{tabular}{|c|>{\centering}p{1.5cm}|>{\centering}p{1.5cm}|>{\centering}p{1.5cm}|>{\centering}p{1.5cm}|}
\hline 
shot & accuracy old digits & accuracy new digits & accuracy only new digits & overall accuracy\tabularnewline
\hline 
\hline 
1 & 0.485 & 0.597 & 0.706 & 0.496\tabularnewline
\hline 
2 & 0.296 & 0.669 & 0.716 & 0.357\tabularnewline
\hline 
3 & 0.288 & 0.310 & 0.67 & 0.289\tabularnewline
\hline 
4 & 0.335 & 0.677 & 0.727 & 0.399\tabularnewline
\hline 
5 & 0.387 & 0.683 & 0.811 & 0.443\tabularnewline
\hline 
6 & 0.382 & 0.720 & 0.806 & 0.438\tabularnewline
\hline 
7 & 0.400 & 0.759 & 0.893 & 0.474\tabularnewline
\hline 
8 & 0.352 & 0.845 & 0.900 & 0.442\tabularnewline
\hline 
9 & 0.487 & 0.502 & 0.820 & 0.479\tabularnewline
\hline 
10 & 0.386 & 0.617 & 0.947 & 0.433\tabularnewline
\hline 
\end{tabular}
\par\end{centering}

\caption{\label{tab:Experimental-results-without-fixed}Experimental results
as in \tabref{Experimental-results-few} , but without fixed first
layer.}
\end{table}
\begin{table}
\begin{centering}
\begin{tabular}{|c|>{\centering}p{1.5cm}|>{\centering}p{1.5cm}|>{\centering}p{1.5cm}|>{\centering}p{1.5cm}|}
\hline 
shot & accuracy old digits & accuracy new digits & accuracy only new digits & overall accuracy\tabularnewline
\hline 
\hline 
1 & 0.569 & 0.508 & 0.517 & 0.558\tabularnewline
\hline 
2 & 0.672 & 0.642 & 0.659 & 0.661\tabularnewline
\hline 
3 & 0.565 & 0.505 & 0.517 & 0.557\tabularnewline
\hline 
4 & 0.649 & 0.501 & 0.529 & 0.626\tabularnewline
\hline 
5 & 0.564 & 0.642 & 0.661 & 0.584\tabularnewline
\hline 
6 & 0.684 & 0.724 & 0.733 & 0.696\tabularnewline
\hline 
7 & 0.364 & 0.515 & 0.517 & 0.398\tabularnewline
\hline 
8 & 0.458 & 0.525 & 0.536 & 0.474\tabularnewline
\hline 
9 & 0.454 & 0.532 & 0.541 & 0.477\tabularnewline
\hline 
10 & 0.451 & 0.784 & 0.792 & 0.508\tabularnewline
\hline 
\end{tabular}
\par\end{centering}

\caption{\label{tab:Experimental-results-without-enriched}Experimental results
as in \tabref{Experimental-results-few} , but without enriched samples.}
\end{table}

\section{Discussion}

Two simple ideas lead to successful few-shot learning: disabling learning
of the first layer during few-shot learning and adding samples from
pre-training to every shot. Both measures could be integrated into
a unified learning procedure. 

For example by reducing the learning rate of the first layer by a
factor of 0.01 the same neural network could be used for pre-training
as well as for few-shot learning. Even if it performs a little worse
than our original procedure, the few-shot learning succeeds (table
\ref{tab:Experimental-results-with-reduced-learning-rate-0.01}).
\begin{table}
\begin{centering}
\begin{tabular}{|c|>{\centering}p{1.5cm}|>{\centering}p{1.5cm}|>{\centering}p{1.5cm}|>{\centering}p{1.5cm}|}
\hline 
shot & accuracy old digits & accuracy new digits & accuracy only new digits & overall accuracy\tabularnewline
\hline 
\hline 
1 & 0.896 & 0.409 & 0.645 & 0.809\tabularnewline
\hline 
2 & 0.754 & 0.573 & 0.699 & 0.722\tabularnewline
\hline 
3 & 0.926 & 0.484 & 0.667 & 0.841\tabularnewline
\hline 
4 & 0.855 & 0.637 & 0.832 & 0.816\tabularnewline
\hline 
5 & 0.836 & 0.730 & 0.821 & 0.813\tabularnewline
\hline 
6 & 0.890 & 0.574 & 0.688 & 0.830\tabularnewline
\hline 
7 & 0.935 & 0.683 & 0.893 & 0.880\tabularnewline
\hline 
8 & 0.879 & 0.778 & 0.881 & 0.860\tabularnewline
\hline 
9 & 0.878 & 0.806 & 0.914 & 0.865\tabularnewline
\hline 
10 & 0.874 & 0.801 & 0.906 & 0.862\tabularnewline
\hline 
\end{tabular}
\par\end{centering}

\caption{\label{tab:Experimental-results-with-reduced-learning-rate-0.01}Experimental
results with pre-training and few-shot learning using the same neural
network with a reduced learning rate of the first layer (by a factor
0.01) .}
\end{table}
\begin{table}
\begin{centering}
\begin{tabular}{|c|>{\centering}p{1.5cm}|>{\centering}p{1.5cm}|>{\centering}p{1.5cm}|>{\centering}p{1.5cm}|}
\hline 
shot & accuracy old digits & accuracy new digits & accuracy only new digits & overall accuracy\tabularnewline
\hline 
\hline 
1 & 0.786 & 0.411 & 0.827 & 0.700\tabularnewline
\hline 
2 & 0.719 & 0.519 & 0.753 & 0.676\tabularnewline
\hline 
3 & 0.793 & 0.589 & 0.880 & 0.745\tabularnewline
\hline 
4 & 0.755 & 0.630 & 0.873 & 0.723\tabularnewline
\hline 
5 & 0.761 & 0.657 & 0.882 & 0.732\tabularnewline
\hline 
6 & 0.771 & 0.651 & 0.884 & 0.738\tabularnewline
\hline 
7 & 0.778 & 0.645 & 0.883 & 0.744\tabularnewline
\hline 
8 & 0.772 & 0.674 & 0.852 & 0.739\tabularnewline
\hline 
9 & 0.765 & 0.734 & 0.888 & 0.751\tabularnewline
\hline 
10 & 0.792 & 0.714 & 0.888 & 0.768\tabularnewline
\hline 
100 & 0.840 & 0.856 & 0.963 & 0.830\tabularnewline
\hline 
\end{tabular}
\par\end{centering}

\caption{\label{tab:The-pre-training-with-few-shot-procedure}The few-shot
learning results are shown, after the pre-training is done with the
few-shot procedure using 5000 shots (mini-batches consisting of one
new sample together with the last four used samples), reaching an
accuracy of about 85\%. This way pre-training and few-shot learning
is done the same way. The neural network is the same one used in previous
experiments, but the learning rate of the first two layers is decreased
by a factor of 0.5 and 0.25 respectively. The first layer is random
initialized with a standard derivative of 0.1, the other layers with
0.5.}
\end{table}

Enriching the few-shot samples with ``old samples'' can be done
straight forward. By keeping one or more samples for each label and
use them as ``old samples'', it is even possible to use the few-shot
learning procedure for pre-training. To test this approach we use
the few-shot procedure with shots of one digit together with the last
four digits as pre-training followed the same few-shot learning as
before (table \ref{tab:The-pre-training-with-few-shot-procedure}).
This way the same procedure is used to learn 8 digits as pre-training,
followed by learning the remaining two digits as few-shot learning.
Using 5000 shots for pre-training, reaching about 85\% accuracy, and
10 shots for few-shot learning we reach more than 75\% over all accuracy.
Using 100 shots for few-shot learning increases the accuracy to 83\%.
Further tests indicate, that increasing the number of shots for pre-training
increases the few-shot accuracy.

\section{Outlook}

The experience from the relatively small problem of hand written digits
indicate, that our few-shot procedure is successful. Some tests increasing
the size of the hidden layers and the amount of pre-training shows
even higher accuracy. Overfitting seems not to be a problem, at least
not for hidden sizes up to 1024. Therefore we expect increasing the
size of the neural network and increasing the number of labels might
increase the accuracy, which would support the use in real world application.

\bibliographystyle{unsrt}
\bibliography{used.bib}

\end{document}